\documentclass[runningheads]{llncs}

\usepackage{graphicx}
\usepackage{amsfonts}
\usepackage{listings}
\usepackage{xspace}
\usepackage{multirow}
\usepackage{array} 
\usepackage{enumitem}
\usepackage[table,xcdraw]{xcolor}
\usepackage{url}
\usepackage{hyperref} 
\usepackage{mathtools}
\usepackage{patchcmd}
\usepackage{multirow,tabularx}
\usepackage{array}
\usepackage{subcaption}

\DeclareGraphicsExtensions{.pdf,.png,.jpg}

\newcommand{\algoname}{Ariadne}

\newcommand{\etal}{\textit{et al. }}
\newcommand{\eg}{e.g. }
\newcommand{\ie}{i.e. }

\DeclareMathOperator*{\argmax}{arg\,max}

\begin{document}

\title{\LARGE \bf Let's take a Walk on Superpixels Graphs: Deformable Linear Objects Segmentation and Model Estimation} 
\titlerunning{Let's take a Walk on Superpixels Graphs} 


\author{Daniele De Gregorio\inst{1}\orcidID{0000-0001-8203-9176} \and
Gianluca Palli\inst{2}\orcidID{0000-0001-9457-4643} \and
Luigi Di Stefano\inst{1}\orcidID{0000-0001-6014-6421}}
%

\authorrunning{De Gregorio et al.} 


\institute{DISI, University of Bologna, 40136
      Bologna, Italy \and
DEI, University of Bologna, 40136
      Bologna, Italy 
}

\maketitle

\begin{figure*}
    \centering
    \includegraphics[width=1\textwidth]{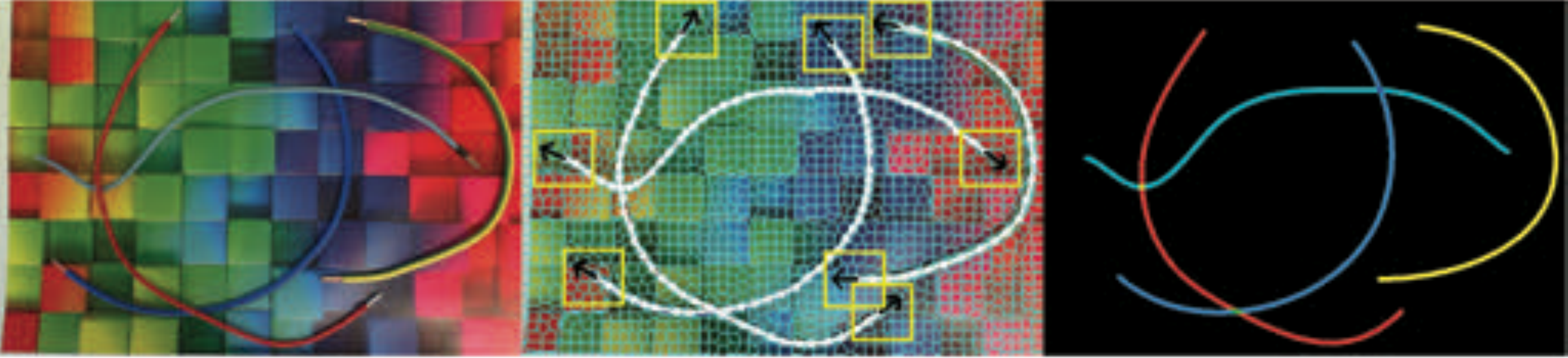}
    \caption{Application of our algorithm to an image featuring a complex background (left). The first kind of output (center) is a \emph{Walk}  (white dots) over the  Region Adjacency Graph (RAG) of the superpixel segmentation which allows for computing a b-spline model of each object. Yellow boxes highlight detection of cables terminals. The second output (right) consists in a segmentation of the whole image.}
    \label{fig:teaser}
\end{figure*}

\vspace{-0.5cm}
\begin{abstract}
While robotic manipulation of rigid objects is quite straightforward, coping with deformable objects is an open issue. More specifically, tasks like tying a knot, wiring a connector or even surgical suturing deal with the domain of Deformable Linear Objects (DLOs). In particular the detection of a DLO is a non-trivial problem especially under clutter and occlusions (as well as self-occlusions). The pose estimation of a DLO results into the identification of its parameters related to a designed model, e.g. a basis spline. It follows that the stand-alone segmentation of a DLO might not be sufficient to conduct a full manipulation task. This is why we propose a novel framework able to perform both a semantic segmentation and b-spline modeling of multiple deformable linear objects simultaneously without strict requirements about environment (i.e. the background). The core algorithm is based on biased random walks over the Region Adiacency Graph built on a superpixel oversegmentation of the source image. The algorithm is initialized by a Convolutional Neural Networks that detects the DLO's endcaps. An open source implementation of the proposed approach is also provided to easy the reproduction of the whole detection pipeline along with a novel cables dataset in order to encourage further experiments.

\end{abstract}
\section{Introduction} \label{sec:introduction}

Plenty of manipulation tasks deal with objects that can be modeled as non-rigid \emph{linear} -- or more generally \emph{tubular} -- structures. In case the task has to be executed by a robot in an unstructured environment, particular effort must be devoted to effectiveness, reliability and efficiency of the automated perception sub-system. 
Tying knots, for example, is a common though hard to automate  activity. In particular, in surgical operations like suturing,  \emph{grasp and knot-tying} is a very important and repetitive sub-task  (\cite{javdani2011modeling}\cite{jackson2015automatic}\cite{saha2007manipulation}).  Similar knot-tying and path planning procedures, like e.g. knots untangling, are also relevant to contexts like service, collaborative and rescue robotics (\cite{lui2013tangled}\cite{nair2017combining}\cite{schulman2013tracking}\cite{hopcroft1991case}). As for industrial scenarios, one of the hardest tasks involving flexible linear objects is  wire routing  in  assembly processes   (\cite{remde1999picking}\cite{yue2002manipulating}\cite{alvarez2016approach}\cite{koo2008development}). 
This paper  focuses primarily on the industrial field and extends the original concepts introduced in the WIRES\footnote{This work was supported by the European Commissions Seventh Framework Programme (FP7/2007-2013) under grant agreement no. 601116.} project and described in \cite{degregorio2018}. As the project aims at automatizing the switchgear wiring procedure, cable modeling by perception is key to address  sub-tasks like  cable grasp, terminal insertion, wireways routing and, not least in importance,  simulation and validation. 
 
In this paper we propose a novel computer vision algorithm for generic Deformable Linear Object (DLO) detection. As highlighted in \autoref{fig:teaser}, the proposed algorithm yields a twofold representation of detected objects, namely a b-spline model for each target alongside with segmentation of the whole image. This twofold representation helps addressing both relatively simpler application settings dealing with cable detection as well as more complex endeavours calling for estimation of cable bend. The proposed algorithm  consists of two distinct modules: the first, which may be  considered as a pre-processing stage, detects  the end-caps regions of DLOs by exploiting off-the-shelf Convolutional Neural Networks (\cite{redmon2017yolo9000,huang2016speed}); the second module, instead, is  the core of this work and allows for identifying DLOs based on the \emph{coarse} position of their endpoints in images featuring complex backgrounds as well as occlusions (e.g. cables crossing other cables) and self-occlusions (e.g. a cable crossing itself, also several times).

The algorithm exploits an over-segmentation of the source image into superpixels to build a Region Adjacency Graph (\emph{RAG} \cite{tremeau2000regions}). This representation enables to detect the area enclosing  each  target object by efficiently analyzing  meaningful regions (i.e. superpixels) only rather than the whole pixels set. The task is accomplished by an iterative procedure capable to find the best path (or  \emph{walk}) through the RAG between two seed points by analyzing several local and global features  (e.g. visual similarity, overall curvature etc.). This iterative procedure yields a directed graph of superpixels conducive to  vectorization and  b-spline approximation.  As better explained in the remainder of the paper, our approach is mostly unsupervised (i.e. only the endpoints detection CNN is trained supervisedly) and relies on just  a few parameters that can be easily tuned manually (or  estimated based on the characteristics of object's material, e.g. elasticity or the plasticity). As we shall see in \autoref{sec:related_works},  our algorithm outperforms other known approaches that one may  apply to try solving the addressed task. 

\section{Related Work}\label{sec:related_works}

\vspace{-1cm}
\begin{table}[]
\centering
\caption{Curvilinear objects alongside with their key features.  \emph{Curvature} expresses the high-level shape representation for that object. \emph{Intersections} indicates whether a crossing between two objects is allowed  or not (\emph{A = Allowed}, \emph{N = Not}); \emph{Bifurcations} denotes whether an object may bifurcate into multiple parts (\emph{A}) or not (\emph{N}). }
\vspace{0.5cm}
\label{table:competitors}
\setlength{\tabcolsep}{0.2cm}
\begin{tabular}{|c|c|c|c|}
\hline
\rowcolor[HTML]{EFEFEF} 
{\color[HTML]{333333} Object Type} & {\color[HTML]{333333} Curvature} & {\color[HTML]{333333} Intersections} & {\color[HTML]{333333} Bifurcations} \\ \hline
\textbf{Cables}                    & \textbf{Spline Model}            & \textbf{A}                           & \textbf{N}                          \\ \hline
Fingerprints                       & Low and Bounded                  & N                                    & A                                   \\ \hline
\textbf{Guidewires}                & \textbf{Spline Model}            & \textbf{N}                           & \textbf{N}                          \\ \hline
Pavement Cracks                    & Random                           & A                                    & A                                   \\ \hline
\textbf{Power Lines}               & \textbf{Spline Model}            & \textbf{A}                           & \textbf{N}                          \\ \hline
Roads                              & Model Based                      & A                                    & A                                   \\ \hline
\textbf{Ropes}                     & \textbf{Spline Model}            & \textbf{A}                           & \textbf{N}                          \\ \hline
\textbf{Surgery Threads}           & \textbf{Spline Model}            & \textbf{A}                           & \textbf{N}                          \\ \hline
Vessels                            & Low and Bounded                  & N                                    & A                                   \\ \hline
\end{tabular}
\end{table}


Although the literature concerning  DLOs  is more focused on manipulation than on perception (see e.g. \cite{saha2007manipulation}), we may refer to the broader topic of  \emph{Curvilinear Objects Segmentation} to highlight related works as well as suitable alternatives  to evaluate our proposal comparatively. As described in \cite{bibiloni2016survey}, the aforementioned topic pertains several kinds of objects, as summarized in \autoref{table:competitors}. Our target objects are \emph{Cables}, which, however, share similar properties in terms of \emph{Intersections} and \emph{Bifurcations} with the other categories highlighted in bold in \autoref{table:competitors},  

As far as \emph{Cables} are concerned, visual perception is typically addressed in fairly simple settings. In  \cite{jiang2011robotized}  \emph{Augmented Reality} markers are deployed  to track end-points. In other works, like \cite{remde1999picking} and \cite{camarillo2008vision}, detection relies on  background removal  \footnote{In \cite{camarillo2008vision} the authors deal with a thin flexible manipulator which may be described as a cable due to the high number of degrees of freedom.}.  

Moving to the domain of  knot-tying with \emph{Ropes},  the basic approach still turns out to be background removal, like in \cite{hopcroft1991case}, or  its  3D counterpart  -- plane removal--  like in \cite{lui2013tangled} and \cite{schulman2016learning}. All these methods produce a raw set of points on which a region growing algorithm is run to attain a vectorization of the target  object. A different approach is used in  \cite{schulman2013tracking}, with the model of the object  registered to the 3D point cloud in real-time in order to avoid the segmentation step. As described in \cite{nair2017combining},  deep learning can also be used to track a deformable object: deep features are  associated with rope configurations so to establish a direct mapping toward energy configurations without any explicit modeling step. This approach, however, may hardly work effectively in presence of complex and/or unknown backgrounds.

In the medical field,  \emph{Surgery Threads} detection is just the same kind of problem, albeit at a smaller scale. Also the literature dealing with this domain is  more focused on manipulation  than on detection issues, either assuming the latter as  solved  \cite{moll2006path} or addressing it by hand-labeled markers  \cite{javdani2011modeling}. A more scalable approach is proposed in  \cite{padoy20113d} and \cite{jackson2015automatic}, where the authors borrow the popular Frangi Filter \cite{frangi1998multiscale} from the field of \emph{vessels segmentation} in order to enhance the curvilinear structure of suture threads and produce a binary segmentation amenable  to estimate a spline model.

Despite  \autoref{table:competitors} would suggest  \emph{Ropes}, \emph{Guidewires}, \emph{Powelines} and \emph{Surgery Threads} to exhibit more commonalities with  \emph{Cables},  applications domains  like \emph{Vessels Segmentation} or \emph{Road detection} can provide interesting insights and solutions for curvilinear object detection. Akin to most object detection problems, the most successful approaches to  \emph{Vessels Segmentation}  leverage on deep learning.  In \cite{liskowski2016segmenting} the authors trained a Convolutional Neural Network (CNN) with hundreds of thousands labeled images in order to obtain a very effective detector.  This supervised approach, however,  mandates availability of a huge training set and lots of man-hours, a combination quite unlikely amenable to real-world industrial settings. Similar considerations apply to other remarkably effective \emph{vessels segmentation} approaches based on supervised learning like \cite{melinvsvcak2015retinal} and \cite{li2016cross}. Yet, in this realm several methods exploiting 2D filtering procedures are very popular and may be applied within a \emph{cable} detection pipeline for industrial applications. The first interesting approach was developed by Frangi \emph{et.al} \cite{frangi1998multiscale} (hereinafter \emph{Frangi} algorithm\footnote{\url{https://github.com/ntnu-bioopt/libfrangi}}) and consists in a multi-scale filtering procedure capable of enhancing tubular structures.  Another methods, described in \cite{staal2004ridge}, deploys a pre-processing stage based on \emph{ridge} detection (hereinafter \emph{Ridge} algorithm\footnote{\url{https://github.com/kapcom01/Curviliniar_Detector}}) to detect vessels. Although the authors use the outcome of this stage to feed a pixel-wise classifier,  we found that the stage itself is very useful to highlight generic tubular structures and hence choose  to compare this algorithm with ours in the experiments. Given that both the \emph{Frangi} and \emph{Ridge} algorithms deal with detection/enhancement of 2D curvilinear structure, we decided to include in the evaluation  also \emph{ELSD}  \cite{puatruaucean2012parameterless}, which is a popular  parameter-less algorithm aimed at detection of line segments and elliptical arcs.

\section{Algorithm Description}\label{sec:algorithm_description}

\begin{figure*}[t] 
  \centering
  \includegraphics[width=0.8\textwidth]{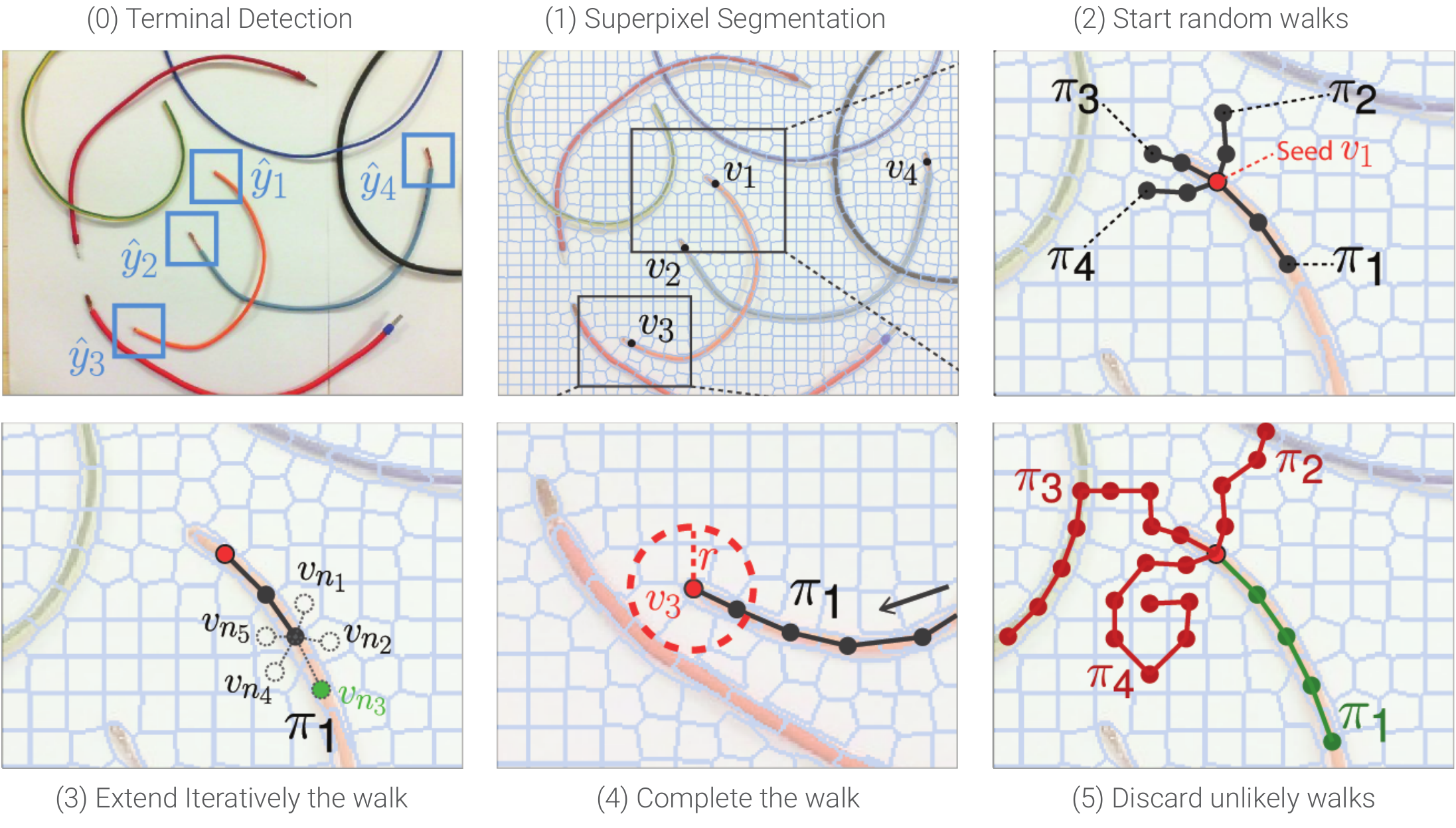}
  \caption{A graphical representation of the whole pipeline described in detail in \autoref{sec:algorithm_description} 
}  \label{fig:pipeline}
 \vspace{-0.5cm}  
\end{figure*}

The basic idea underlying our algorithm is to detect DLOs as suitable \emph{walks} within an adjacency graph built on superpixels. We provide first an overview of the approach with the help of \autoref{fig:pipeline}, which illustrates the following main steps of the whole pipeline. 

\begin{enumerate}[start=0]
    \item \textbf{Endpoints Detection}: The first step consists in detecting endpoints. This is numbered as zero because it may be thought of as an external process not tightly linked to the  rest of the algorithm. Indeed, any external algorithm capable to produce detections around targets ($y_1 ... y_k$) may be deployed in this step.
   
   \item \textbf{Superpixel Segmentation}: The source image $\mathcal{I}$ is segmented into adjacent sub-regions (superpixels) in order to build a set of segments exhibiting a far smaller cardinality than the whole pixel set. Moreover, an adjacency graph is created on top of this segmentation in order to keep track of the neighborhood of each superpixel. Eventually, those superpixels containing the 2D detections obtained in  the previous step are marked as \emph{seeds} ($v_1 ... v_k$).
   
   \item \textbf{Start Walks}: From each  \emph{seed} we can start an arbitrary number of \emph{walks} ($\pi_1 ... \pi_P$)  by moving into adjacent superpixels as defined by the adjacency graph.  
   
   \item \textbf{Extend Walks}: For each \emph{walk} we  move forward along the adjacency graph by choosing iteratively the best next superpixel (\eg $v_{n_3}$ in \autoref{fig:pipeline}-(3)) between the neighbourhood  \{$v_{n_1}, ..., v_{n_c}$\} of the current one. 
   
   \item \textbf{Terminate Walks}: When a \emph{walk} reaches another \emph{seed} (or lies in its neighborhood) it is  marked as closed. Due to the iterative nature of the computation, a maximum number of extending steps is allowed for each search to ensure a bounded time complexity.
   
   \item \textbf{Discard Unlikely Walks}: As a set of random walks are started in \emph{Step 2}, we keep only the most likely  ones and mark others as outliers.
   
\end{enumerate}

\begin{figure}[]
\centering
  \begin{tabular}{c c c}
  \multicolumn{3}{c}{\includegraphics[width=0.8\textwidth]{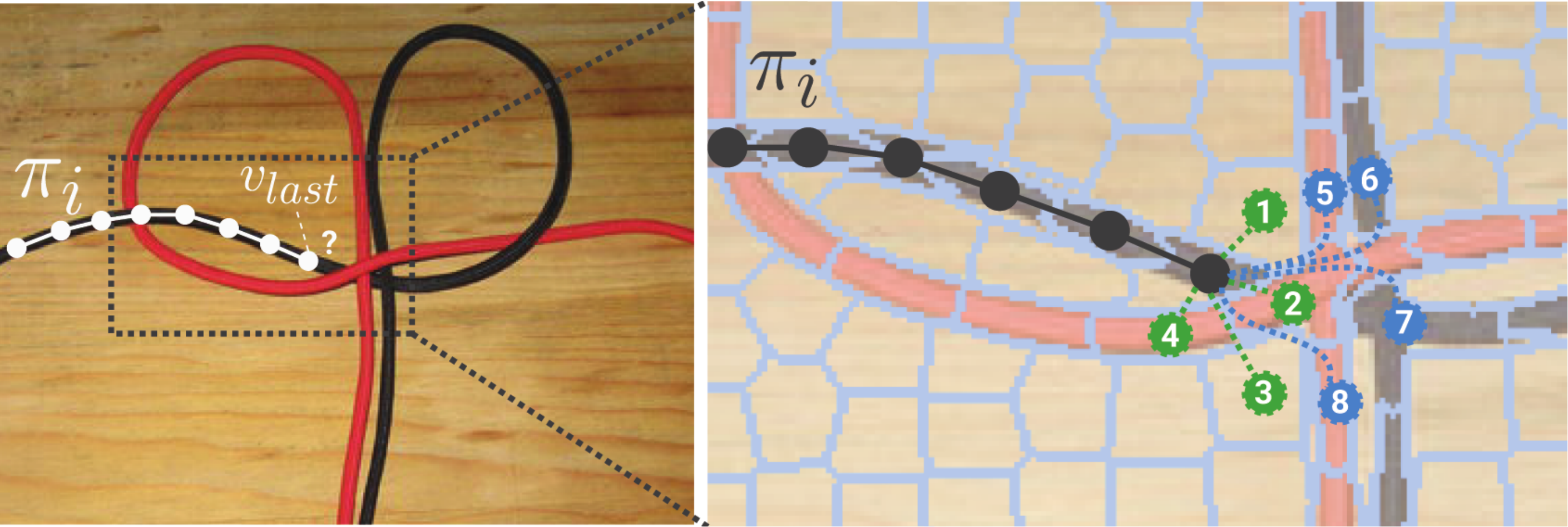}} \\
  \multicolumn{3}{c}{(1) Example of crossing and self-crossing wires } \\
  \multicolumn{3}{c}{ } \\
  \includegraphics[width=0.25\textwidth]{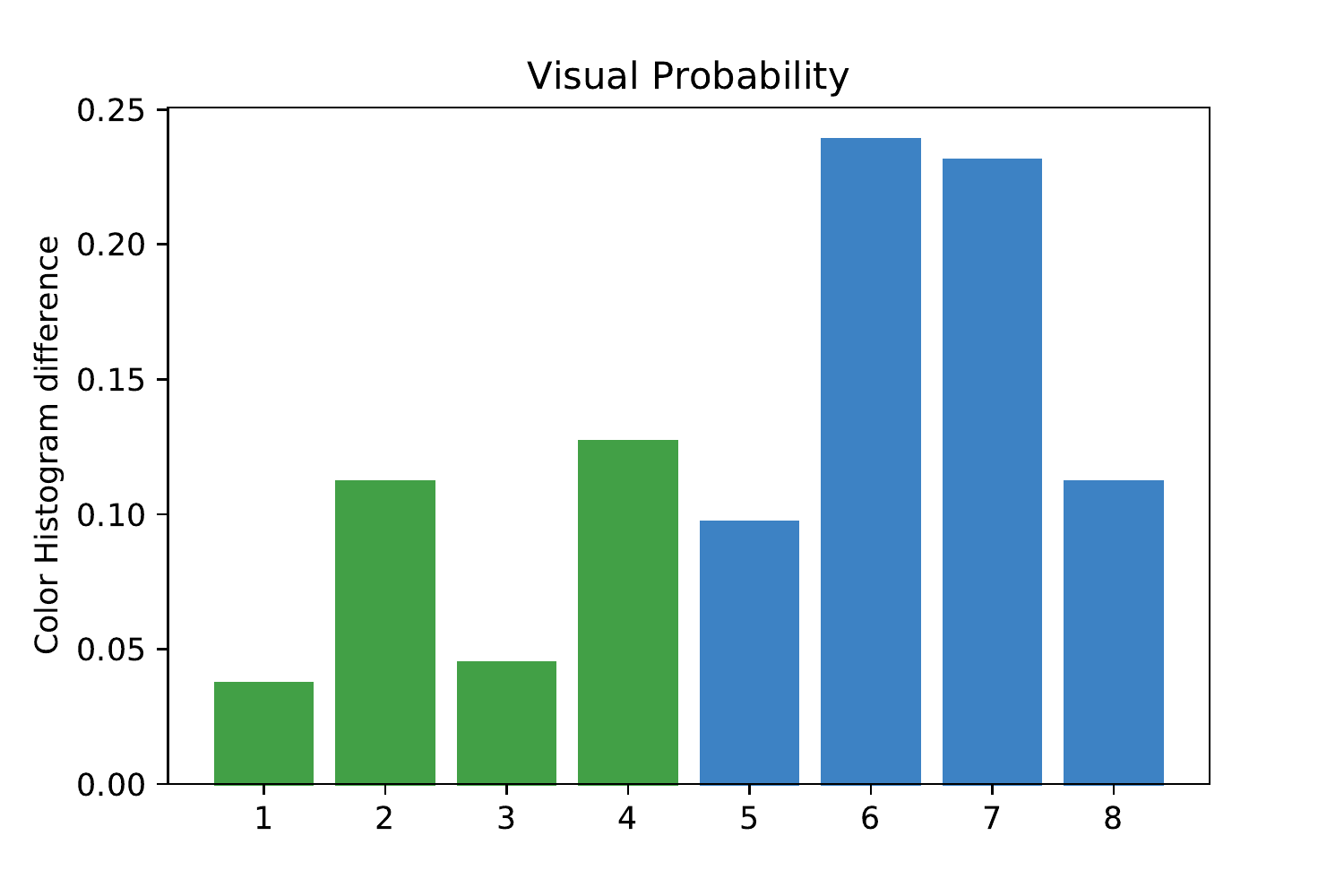} &
  \includegraphics[width=0.25\textwidth]{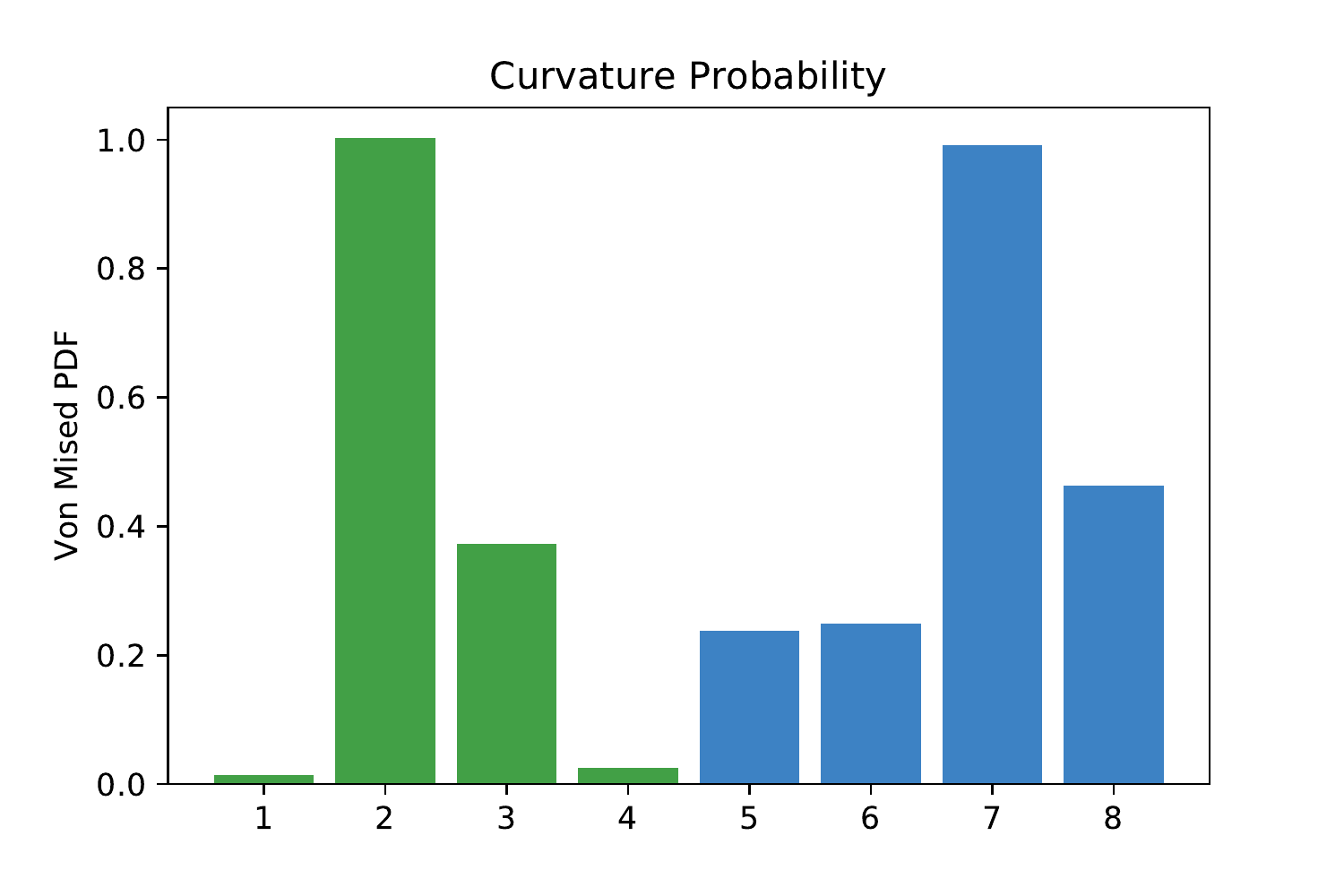} &
  \includegraphics[width=0.25\textwidth]{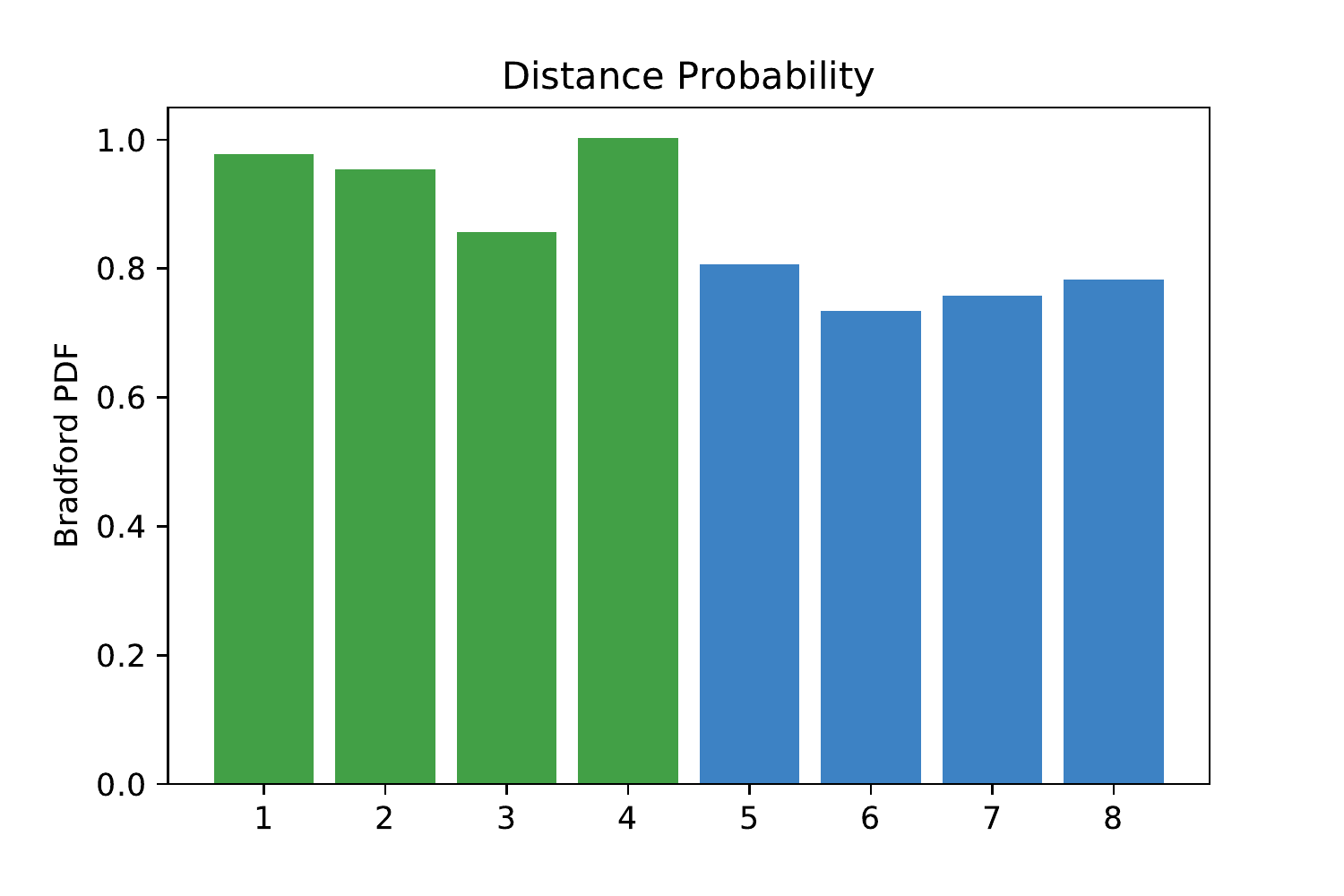} \\
  (2)~Visual likelihood & 
  (3)~Curvature likelihood & 
  (4)~Distance likelihood
  \end{tabular}
  \caption{(1) shows a complex configuration dealing with crossing and self-crossing wires. The zoom highlights the last node ($v_{last}$) alongside with the candidates to select the next node, \emph{i.e.} first and second order neighbours (green and blue dots, respectively). (2),(3) and (4) plot the Visual, Curvature and Distance likelihoods, respectively. Based on the contribution of all the three likelihood terms,  node $7$ is selected to extend the current \emph{walk}.}
  \vspace{-0.5cm}
  
  \label{fig:walk_probability}
\end{figure}

In the remainder of this Section we will describe in detail the different concepts, methods and computations needed to realize the whole pipeline. In particular, in \autoref{sec:superpixel_segmentation} we address  Superpixels together with the Region Adjacency Graph; in \autoref{sec:walk_construction} we define  \emph{walks} and how they can be  built iteratively  by analyzing local and global features; in \autoref{sec:random_paths} we propose a method to start the random walks by exploiting an external  Object Detector based on a CNN and, to conclude, in \autoref{sec:semantic_segmentation} we describe how to deploy walks to attain a semantic segmentation of the image as depicted in \autoref{fig:teaser}.

\subsection{Superpixel Segmentation and Adjacency Graph}
\label{sec:superpixel_segmentation}

The main aim of  \emph{Superpixel Segmentation} is to replace the rigid structure of the pixel grid with an higher-level subdivision into more meaningful primitives called \emph{superpixels}. These primitives group regions of perceptually similar raw pixels, thereby potentially reducing the computational complexity of further  processing steps. As regards the Cable Detection and Segmentation problem (\ie DLOs with a thickness higher than  a single pixel, in the majority of  applications), our assumptions are that the target wire can be represented as a subset of \emph{similar} adjacent superpixels. Thus, the overall problem can be seen as a simple iterative search through the superpixels set subject to model-driven constraints (\eg avoiding solutions with implausible curvature or  non-uniform visual appearance).

\begin{figure*}
\centering
  \begin{tabular}{c c c c}
  
  \includegraphics[width=0.2\textwidth]{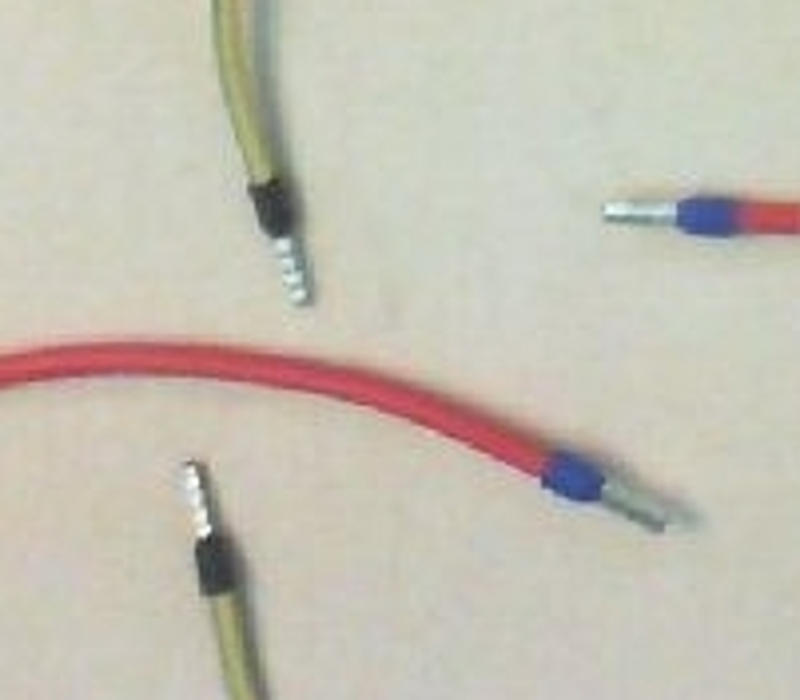} &
 \includegraphics[width=0.2\textwidth]{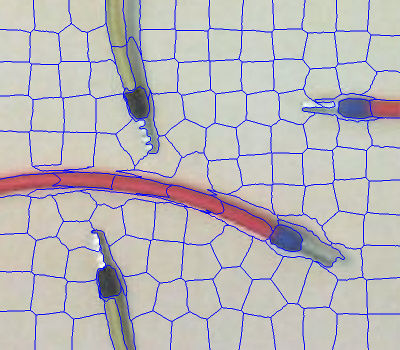} &
 \includegraphics[width=0.2\textwidth]{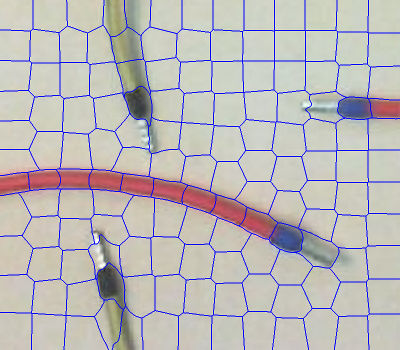} &
 \includegraphics[width=0.2\textwidth]{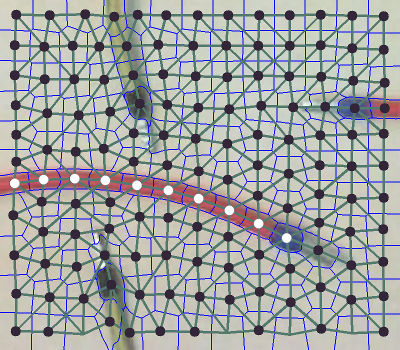} \\
  (a) &
  (b) &
  (c) &
  (d) 
  \end{tabular}
  \caption{(a) Original input image. (b) SLIC superpixels segmentation with low compactness. (c) Segmentation with high compactness. (d) The Region Adjacency Graph (RAG) built on the (c) segmentation.}
  \vspace{-0.5cm}
  \label{fig:segmentations}
\end{figure*}

\emph{Superpixel Segmentation}  algorithms can be categorized into graph-based approaches (\eg the method proposed by  Felzenszwalb \etal\cite{felzenszwalb2004efficient}) and gradient-ascent methods (\eg \emph{Quick Shift} \cite{vedaldi2008quick}). In our experiments we found  the state-of-the-art algorithm referred to as \emph{SLIC} \cite{achanta2012slic} to perform particularly well in terms of both speed and accuracy. Accordingly, we deploy  \emph{SLIC} in our \emph{Superpixel Segmentation} stage. \emph{SLIC} is an adaptation of the \emph{k}-means clustering algorithm  to pixels represented as 5D vectors  $(l_i, a_i, b_i, x_i, y_i)$, with $(l_i, a_i, b_i)$ denoting  color channels in the CIELAB space and $(x_i, y_i)$ image coordinates. During the clustering process the compactness of each cluster can be either increased or reduced to the detriment of  visual similarity. In other words we can choose easily to assign more importance to visual consistency of superpixels or to their spatial uniformity.  \autoref{fig:segmentations} (b),(c) show  two segmentations provided by \emph{SLIC} according to different settings for the  visual consistenvy vs. spatial uniformity trade-off.   

 \emph{Superpixel Segmentation} allows then to  build  a Region Adjacency Graph (RAG in short)  according to the method described in \cite{tremeau2000regions}. Thus, a generic image $I$ can be partitioned into disjoint non-empty regions $R_0, ..., R_i$ (\ie the superpixels) such as {\small$I = \bigcup R_i$}. 
Accordingly, an undirected weighted graph si given by $G=(V,E)$, where $V$ is the set of nodes $v_i$,  corresponding to each region $R_i$, and $E$ is the set of edges $e_{j,k}$ such as that $e_{j,k} \in E$ if $R_i$ and $R_j$ are adjacent.
A graphical representation of this kind of graph is shown in  \autoref{fig:segmentations}-(d), where black dots represents  nodes $v_i$ and black lines represent edges $e_{j,k}$. In this quite straightforward to observe that in \autoref{fig:segmentations}-(d) there  exists a \emph{walk} trough the graph $G$, highlighted by white dots, which covers a  target DLO  (\ie the red cable in the middle). 

It is worth pointing out that our approach is similar to a Region Growing algorithm, with a seed point corresponding to the cable's tip and  the search space bounded by the RAG. The main difference with a classical Region Growing approach is that we restrict the search along a \emph{walk} applying several model-base constraints rather than relying only on visual similarity only. In particular, the shape of the \emph{walk} is considered by assigning geometric primitives to the elements of the adjacency graph, \ie 2D points for our nodes $v_i$ and 2D segments for our edges $e_{j,k}$, as further described in \autoref{sec:walk_construction}. In simple terms, the geometric consistency of the curve superimposed on a \emph{walk} is analyzed to choose the next node during the iterative search, and all unlikely configurations are discarded. 

\subsection{Walking on the Adjacency Graph}\label{sec:walk_construction}  

Formally, a walk over a graph $G=(V,E)$ is a sequence of alternating vertices and edges  $(v_{i_1}, e_{{i_1},{i_2}}, v_{i_2}, ..., v_{i_{l-1}}, e_{i_{l-1},i_{l}}, v_{i_l} )$, where an edge $e_{i_j,i_k}$ connects  nodes $v_{i_j}$ and $v_{i_k}$, and $l$ is the length of the walk. The definition of \emph{walk} is more general with respect to the \emph{path} or \emph{trail} over the graph because it admits repeated vertices, a common situation when dealing with self-crossing cables. It is important to notice that the Region Adjacency Graph shown in \autoref{fig:segmentations}(d) is a \emph{simple-connectivity relationship} graph, or, equivalently,  connectivity is of order $d=1$, with this meaning that only directly connected regions are mapped into the graph. We can build also RAGs with order $d>1$, thereby allowing, for example, second or third order connectivity. All this translates into the possibility to \emph{jump} during the walk also to the vertex not directly connected to the current region. This turns out very useful, for example, to deal with intersections like that depicted in \autoref{fig:walk_probability}, where vertices $v_1,v_2,v_3,v_4$ are of order $d=1$ and vertices $v_5,v_6,v_7,v_8$  of order $d=2$.

For the sake of simplicity, we can define a generic \emph{walk} as $\pi_i = (v_{i_1},...,v_{i_l})$, \ie an ordered subset of vertices, without considering edges. Under the hypothesis that the target walk $\pi_i$ is superimposed to a portion of the object, the problem is to \emph{extend} the walk in such a way that the next  node $v_{i_{l+1}}$  does belong to the sought DLO. An exemplar situation is illustrated  in \autoref{fig:walk_probability}, where we have a current path $\pi_i$  which ends with $v_{\text{last}}$ and we wish  to choose between the 8 vertices the best one to extend the \emph{walk}.  Considering the new path $\hat{\pi}_{i,v_n}$, \ie the path $\pi_i$ with the addition of vertex $v_n$, we cast the problem as the estimation of the likelihood of the  new path given the current one, which we denote as $p(\hat{\pi}_{i,v_n}\lvert\pi_i)$. Moreover, we estimate this likelihood based on visual similarity, curvature smoothness and spatial distance features and assume these features to be independent:  

\begin{equation}
\label{eq:next_node_probability}
p(\hat{\pi}_{i,v_n}\lvert\pi_i) =  p_{V}( \hat{\pi}_{i,v_n}\lvert\pi_i) \cdot p_{C}( \hat{\pi}_{i,v_n}\lvert\pi_i) \cdot p_{D}(\hat{\pi}_{i,v_n}\lvert\pi_i)
\end{equation}

\noindent  The three terms  $p_V(\cdot)$, $p_C(\cdot)$ and $p_D(\cdot)$ are referred to as  \emph{Visual}, \emph{Curvature} and  \emph{Distance} likelihood, respectively, and computed as follows. 


\subsubsection{Visual Likelihood}
$ p_{V}( \hat{\pi}_{i,v_n}\lvert\pi_i)$ measures the visual similarity between the previous path $\pi_i$ and the path achievable by adding node $v_n$. Assuming an evenly coloured DLO,  we can compute this similarity by matching only the last node of $\pi_i$, $v_{\text{last}}$, with $v_n$. Although, in principle,  it may be possible to use any arbitrary visual matching function,  as highlighted in  \cite{ning2010interactive} we found the Color Histogram of the superpixels associated with vertices to be  a good feature to compare two image regions.  Denoting  as $H_{last}$ and $H_{n}$ the normalized color histograms (in the HSV color space) of the two regions associated with $v_{last}$ and $v_n$, respectively, we can compute their distance with the intersection equation: ${d(H_{last},H_n) = \sum\limits_I min(H_{last}(I), H_n(I))}$. Then we normalize this distance in the range $[0,1]$ using the \emph{Bradford} normal distribution:

\begin{equation}
 p_{V}( \hat{\pi}_{i,v_n}\lvert\pi_i) = \frac{c_V}{log(1+c_V)(1 + c_V(1-d(H_{last},H_n))}
\end{equation}
\label{eq:bradford_visual}

\noindent where $c_V$ is a parameter that enables to control the shape of the distribution and, hence, the weight assigned to the visual similarity information in the overall computation of the likelihood (\autoref{eq:next_node_probability}).  \autoref{fig:walk_probability}(2) plots the visual likelihoods computed for the different neighbours of $v_{last}$, which suggests  nodes $6$ and $7$ to represent  the most likely superpixels to extend the  \emph{walk}.

\subsubsection{Curvature Likelihood}
\begin{figure*}[t] 
  \centering
  \includegraphics[width=0.3\textwidth]{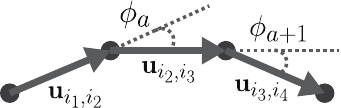}
  \caption{A generic unit vector $\mathbf{u}$ can be assigned to each edge in the adjacency graph so as to compute  the angle difference $\phi_a$, between  consecutive edges. }  \label{fig:vonmises}
\end{figure*}

$p_{C}( \hat{\pi}_{i,v_n}\lvert\pi_i)$ is concerned with estimating the most likely  configuration of a DLO's curvature.  Following the intuitions of Predoehl \etal \cite{predoehl2013statistical}, for each new node $v_n$ we can assume that the object's  curvature changes smoothly  along the \emph{walk}. To quantify  this smoothness criterion we exploit the product of the von Mises distributions of the angles between two successive vertices. As introduced in \autoref{sec:superpixel_segmentation}, by  extending the model of our adjacency graph with geometric primitives we can assign a 2D point $\mathbf{p}_{i_j}$ corresponding to the centroid of the associated superpixel to  each vertex $v_{i_j}$, as well as a unit vector $\mathbf{u}_{i_{j,k}}$ to each edge $e_{i_{j,k}}$ by considering  the segment joining  two consecutive centroids $\mathbf{p}_{i_j}$, $\mathbf{p}_{i_k}$. As shown in \autoref{fig:vonmises}, this allows for measuring the angle difference  between two consecutive edges. By denoting as  $\phi_a =  \phi_{j,k,m}$ the angle difference between two consecutive edges $\mathbf{u}_{i_{j,k}}$,  $\mathbf{u}_{i_{k,m}}$, the overall von Mises  distribution allowing to establish upon the smoothness of the curvature of a target DLO is given by:

\begin{equation}
\label{eq:vonmises_prod}
p_{C}( \hat{\pi}_{i,v_n}\lvert\pi_i) = \prod_a \mathcal{M}(\frac{\phi_a  - \phi_{a+1}}{2},m)
\end{equation}

\noindent where $\mathcal{M}(\cdot)$ is the von Mises distribution at each vertex.  An exemplar estimation is shown in \autoref{fig:walk_probability}(3): vertices $2$ and $7$ appear to be the most likely candidates to extend the \emph{walk} as they minimize the curvature changes of the target $\pi_i$.

\subsubsection{Distance Likelihood}
$p_{D}( \hat{\pi}_{i,v_n}\lvert\pi_i)$ is the term concerned with the spatial distance of the next vertex in the \emph{walk}. This term is mainly introduced to force the iterative procedure to choose the nearest available vertex without undermining the chance to pick a far vertex instead, for example when we want to deal with  an intersection (see \autoref{fig:walk_probability}(1)). Thus, similarly to \autoref{eq:bradford_visual} we normalize  the distance in pixel between two  nodes, $d_{pixel}(\mathbf{p}_{i_{last}},\mathbf{p}_{i_{n}})$, according to the Bradford normal distribution: 

\begin{equation}
p_{D}( \hat{\pi}_{i,v_n}\lvert\pi_i) = \frac{c_D}{log(1+c_D)(1 + c_D(1-d_{pixel}(\mathbf{p}_{i_{last}},\mathbf{p}_{i_{n}})}
\end{equation}
\label{eq:bradford_distance}

\noindent with $c_D$ tuned such that the decay of the distribution is slow to prefer nearest vertex but not enough to discard the furthest points.  \autoref{fig:walk_probability}(4) highlights how, thanks to the normalization in \autoref{eq:bradford_distance}, second order neighbours  ($5, ..., 8$) are not excessively penalized with respect to first order ones  ($1, ..., 4$) and hence  have the chance to be picked in case they exhibit a high visual similarity and/or yield a particularly smooth \emph{walk}.

\subsubsection{Estimation of the most likely walk}
$p(\hat{\pi}_{i,v_n}\lvert\pi_i)$ can therefore be computed for all considered neighbours in order to pick the most likely vertex, $v_n$, to extend the \emph{walk}, with: 

\begin{equation}
n = \argmax_n p(\hat{\pi}_{i,v_n}\lvert\pi_i)
\end{equation}
\label{eq:vertex_argmax}

\noindent Considering again the example in \autoref{fig:walk_probability}, although the farthest from $v_{last}$, vertex  $7$ is selected to extend the \emph{walk} as it shows  a high visual likelihood as well as a high curvature likelihood.

\subsection{Starting and Terminating Walks}\label{sec:random_paths}

As described \autoref{sec:algorithm_description}, \emph{walks} need to be initialized with \emph{seed} superpixels located at DLOs' endpoints. Purposely, we deployed a Convolutional Neural Network to detect endpoints. In particular,  we fine-tuned the publicly available YOLO v2 model \cite{redmon2017yolo9000} pre-trained on ImageNet based on the images from our \emph{Electrical Cable} Dataset (see \autoref{sec:software_dataset}) and by performing several data augmentations. As already mentioned, the endpoint detection module may be seen as an external process with respect to our core algorithm and, as such, in the  comparative experimental evaluation we will use the same set of endpoints obtained by YOLO v2 to initialize all considered methods. 

As illustrated in \autoref{fig:pipeline}(1), the endpoint detection step predicts  a set of bounding boxes  $\hat{y}_i$ around the actual endpoints. For each such prediction we find the superpixel containing the central point of the box. The graph vertex $v_i$ corresponding to this superpixel is marked as a \emph{seed} to start a new \emph{walk}. As no prior information concerning the direction of the best \emph{walk} across the target DLO is available,  multiple \emph{walks} are actually started, in particular along each possible direction (see \autoref{fig:pipeline}(3)). It is worth pointing out that the considered directions are those defined by the \emph{seed} vertex and all  its neighbours in the graph, which, as discussed in \autoref{sec:walk_construction}, can be both first order  ($d=1$)  as well as  higher order ($d>1$) neighbours. Each started  \emph{walk}, then, is iteratively extended according to the  procedure described in \autoref{sec:walk_construction}. .

As for the criteria to terminate \emph{walks}, first of all we set a maximum number of iterations to extend  a \emph{walk}. We also terminate a \emph{walk} if it reaches another \emph{seed} vertex in the adjacency graph. More precisely, as depicted in \autoref{fig:pipeline}(5), we terminate a \emph{walk} if the distance from the current vertex and a \emph{seed} is smaller than a radius threshold $r$.
Thus, given a \emph{seed}, all  \emph{walks} started from that \emph{seed} will terminate and we will have to pick only the optimal one. Purposely, we  exploit again the \emph{Curvature} analysis described in \autoref{sec:walk_construction} and use a formulation similar to \autoref{eq:vonmises_prod}  to pick the smoothest path (\ie the \emph{walk} with the highest value of $P_{C}(\cdot)$).

\subsection{Segmentation and Model Estimation}\label{sec:semantic_segmentation}

The simplest technique to segment the image is to assign different labels to the superpixels belonging to the different \emph{walks} alongside with a background label to those superpixels not included into any \emph{walk}. 
Besides, we estimate a B-Spline approssimation  (with the algorithm described in \cite{dierckx1995curve}) for each walk based on the centroids of the its superpixels $\mathbf{p}_0, ..., \mathbf{p}_l$. Then, given the B-Spline model we can refine the segmentation by building  a pixel mask. In particular, by evaluating the smoothing polynomial we can sample densely the points belonging to each parametrised curve and then adjust the thickness of the segmented output based on the mean size of the superpixels belonging to the \emph{walk}.  The accurate segmentation provided by this procedure is shown  in \autoref{fig:teaser}, where the right image contains the colored B-Splines built over the walks represented in the middle image (the color is estimated by averaging the color of the corresponding superpixels).


\section{Experimental results}\label{sec:experimental_results}

\subsection{Software and Dataset}\label{sec:software_dataset}

We provide an open source software framework called \emph{\algoname{}} available online \footnote{\url{https://github.com/m4nh/ariadne}}. The name \emph{\algoname{}} is inspired by the name of Minos's daughter, who, in Greek mythology, used a thread to lead Theseus outside the Minotaur's maze. The software is written in Python and implements the described approach. 
We created also a novel dataset \footnote{\url{https://github.com/m4nh/cables_dataset}}
which, to the best of our knowledge, is the first \emph{Electrical Cable} dataset  for detection and segmentation. We used this dataset to perform quantitative and qualitative evaluations of our approach with respect to others curvilinear structure detector.

The Dataset is made up of two separated parts: the first consists of 60 cable images with homogeneous backgrounds (white, wood, colored papers, etc.); the second includes  10 cable images with complex backgrounds. In  \autoref{fig:comparisons}, the first 3 rows deal with homogenous backgrounds whilst the other with complex one.  

For each image in the dataset we provide an hand-labeled binary mask superimposed over each target cable separately and an overall mask that is the union of them. Furthermore, we provide a discretization of the B-Spline for each cable in every image which consists in a set of 2D points in pixel coordinates useful to have a lighter model of the cable and track easily its endings. Further details can be found on the dataset website \footnote{\url{https://github.com/m4nh/cables_dataset}}.
.

\subsection{Segmentation results}\label{sec:segmentation_experiments}

\vspace{-0.4cm}
\begin{figure}
	\centering
	\includegraphics[width=1\textwidth]{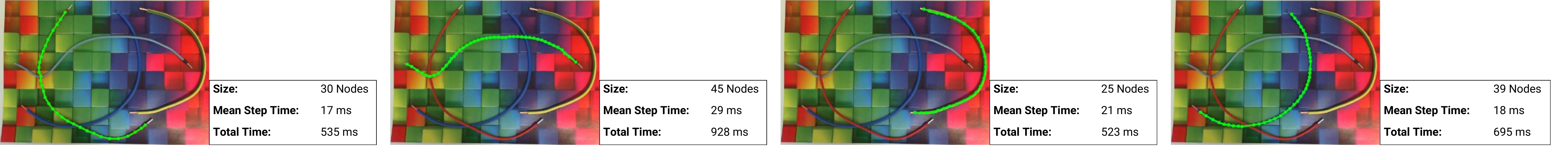}
	\caption{Segmentation timings (complex background).
	}
	\label{fig:timings}
\end{figure}
\vspace{-0.4cm}

To test our approach we compared it to the popular Curvilenar Object Detector discussed  in \autoref{sec:related_works}: the Frangi 2D Filter \cite{frangi1998multiscale}, the Ridge Algorithm \cite{staal2004ridge} and the more generic one ELSD \cite{puatruaucean2012parameterless}. For each algorithm we produce a  mask $M_{prediction}$ associated with the  detected curvilinear structures and compare it, by means of  the Intersection Over Union, $IoU = \frac{M_{gt} \cap M_{prediction} }{ M_{gt} \cup M_{prediction}}$, to the ground truth $M_{gt}$ provided by our dataset. \autoref{table:competitors} reports the weighted  $IoU$ on  $N$ images, where the weight is proportional to the number of cables $C$ present in the image:  $IoU_{weighted} = \frac{1}{C\cdot N}\sum_i C \cdot IoU_{i}$.
The first row refers to images featuring a homogeneous background (60 images with a total of 395 cables), the second to those having a complex background (10 images and a total of 40 cables). We tuned the hyperparameters of the three competing methods  trying to choose the best configuration in order to  cope with both the simpler and harder dataset images. As for our algorithm, we used: a 3D Color Histogram with 8 bins for each channel as Visual similarity feature; a Von Mises distribution with $m=4$ to compute the Curvature likelihood and a degree $d=3$ for the adjacency graph (\ie it means that we search in a neighborhood of level 3 during our walk construction, as described in \autoref{sec:walk_construction}).
For both our approach and the competitors we exploit the information about the endpoints  provided by the inintialization step:  for \emph{Ariadne} we use this information to start random walks, for the competitors to discard many outliers. 

The results reported in \autoref{table:competitors} show  how our approach  outperforms the other methods by a large margin, although it is fair to point out that the competitors are generic curvilinear detectors and not specific DLO detectors. It is also worth highlighting that our method is remarkably robust with respect to complex backgrounds and that this is achieved without any training or fitting procedure that would hinder applicability to unknown scenarios. Thus, our approach  could be used in any real industrial application without requiring prior knowledge of the environment. 

\begin{figure*}[t] 
  \centering
  \includegraphics[width=1\textwidth]{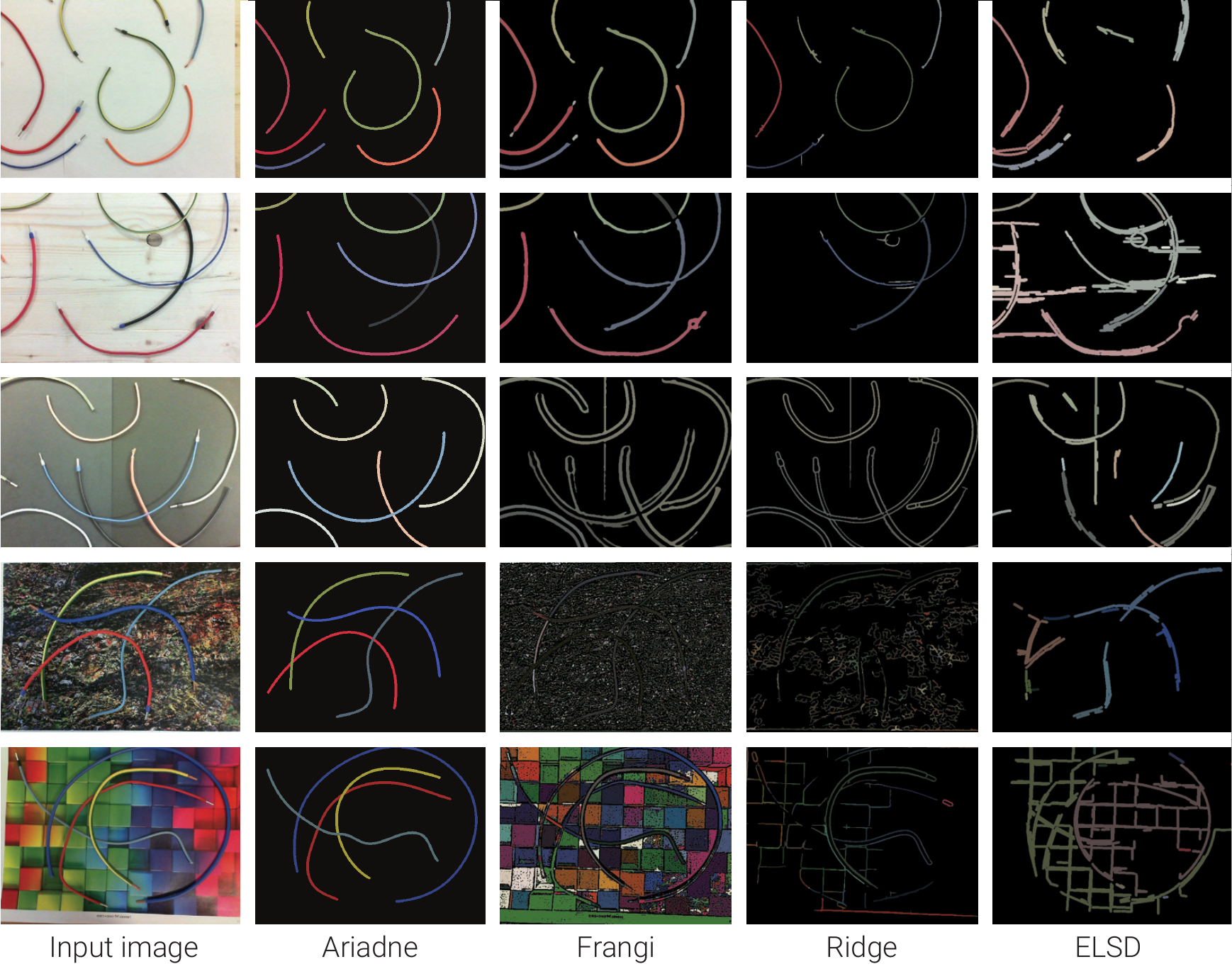}
  \caption{Qualitative evaluation of the different algorithms. The first column represent the input image, whereas the other columns, in order: the result yielded by \emph{Ariadne}, \emph{Frangi}\cite{frangi1998multiscale}, \emph{Ridge}\cite{staal2004ridge} and \emph{ELSD}\cite{puatruaucean2012parameterless}. The difficultly increases from a row to the following: indeed the first row represent a very simple example with white background while the last example, instead, can be considered very hard. Our approach is very robust despite background complexity. }
  \label{fig:comparisons}  
\end{figure*}

Finally,  \autoref{fig:comparisons}, present some qualitative results obtained by our approach. Moreover, an additional qualitative evaluation is present in the supplementary material, where interactive examples of the Ariadne software are shown. In the abovementioned material, as a naive proof-of-concept, we tested Ariadne also in similar challenging contexts like \emph{Roads} and \emph{Rivers} segmentation.

\vspace{-0.5cm}
\begin{table}[]
\large
\centering
\caption{The Intersection Over Union of the cable segmentations obtained with our approach compared with the three major curvilinear structure detectors. }
\vspace{0.2cm}
\label{table:results}
\setlength{\tabcolsep}{5pt}
\setlength{\extrarowheight}{0.5cm}
\begin{tabular}{c|c|c|c|c|}
\cline{2-5}
& \textbf{Ariadne}
&  \textbf{Frangi} \cite{frangi1998multiscale} 
& \textbf{Ridge} \cite{staal2004ridge} 
& \textbf{ELSD} \cite{puatruaucean2012parameterless} 
 \\ \hline
\multicolumn{1}{|l|}{\shortstack{Homegeneous\\Background}}
& \textbf{0.754} 
& 0.406
& 0.293

& 0.225                        
\\ \hline
\multicolumn{1}{|l|}{\shortstack{Complex\\Background}}   
& \textbf{0.583}
& 0.063  
& 0.023                          
                        
& 0.147                        
\\ \hline
\end{tabular}
\end{table}
\vspace{-1cm}

\subsection{Timings and failure cases}

\begin{figure}[htbp]
    \centering
    \begin{subfigure}[b]{0.45\textwidth}
        \includegraphics[width=\textwidth]{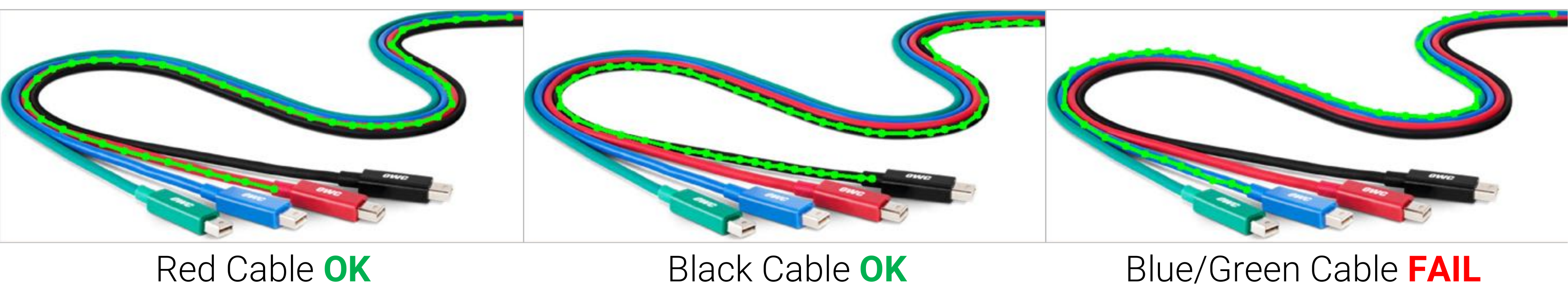}
        \caption{Failure due to adjacent cables.}\label{fig:failure1}
    \end{subfigure}
    \begin{subfigure}[b]{0.45\textwidth}
        \includegraphics[width=\textwidth]{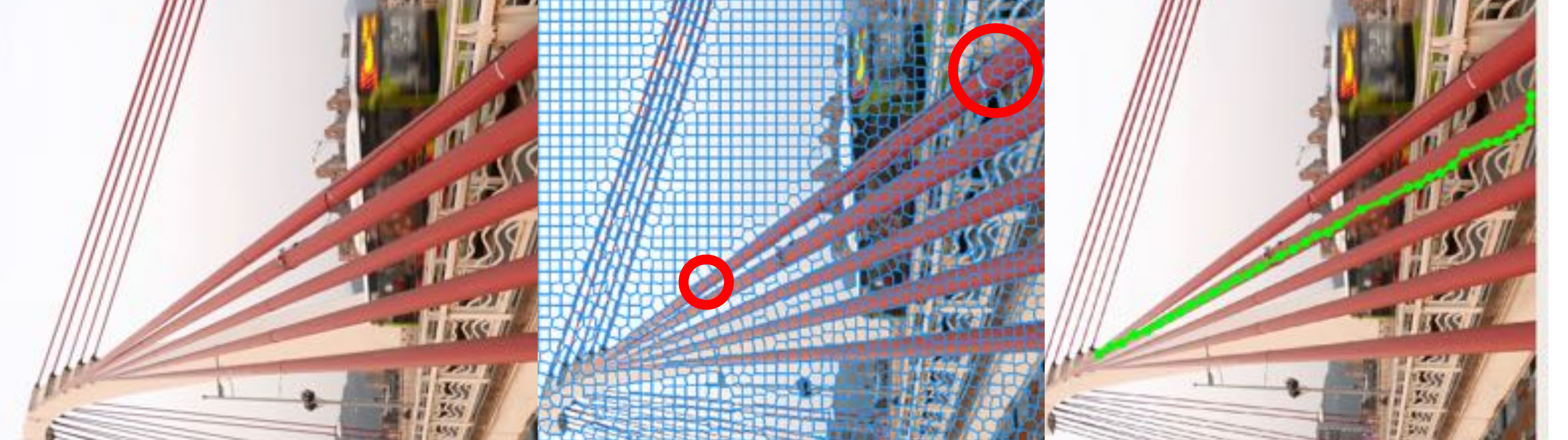}
        \caption{Failure due to perspective.}\label{fig:failure2}      
    \end{subfigure}
     \vspace{-0.3cm}
    \caption{Failure Cases.}
	\label{tab:failures}
    \vspace{-1.5cm}
\end{figure}

\emph{Ariadne} is an iterative approach, with the number of iterations depending on the length of the DLO. Thus, as depicted in \autoref{fig:timings}, we can estimate an average iteration time of about $21ms$ and an average segmentation time of about $670 ms$. We also point out that these measurements were obtained with the actual Python implementation. Moreover, \autoref{tab:failures}(a),(b) shows the two main failure cases that we found for \emph{Ariadne}. In (a), as two DLOs (blue and green) are adjacent  and  exhibit very similar colour and curvature, the walk may jump on the wrong cable. In (b), as the distance between the DLO and the camera varies greatly,  the density of Superpixels is not constant and the walk can cover only a portion of the sought object or even completely fail.

\section{Concluding Remarks}
 
 We presented an effective unsupervised approach to segment DLOs in images. This segmentation method may be deployed in industrial applications involving wire detection and manipulation. Our approach requires an external detector to localize cable terminals, as otherwise  we should start \emph{walks} at every superpixel, which  would be almost unworkable, although not impossible. So far we deploy an external detector which provides only the approximate position of the endpoints. We are currently working to develop  a smarter endpoint  detector capable to infer also the orientation of the cable terminal in order to dramatically shrink  the number of initial \emph{walks}. Another future development concerns  building  a much larger Electrical Cable dataset equipped with ground truth information suitable to train a specific CNN aimed at cable segmentation and compare this supervised approach to \emph{Ariadne}. It is worth pointing out that, in specific and known in advance settings, a supervised approach may turn out peculiarly effective: in such circumstances  \emph{Ariadne} could be used to vastly ease and speed up the manual labeling procedure required to obtain the training images by replacing the initial object detector with the interactive intervention of the user.

\clearpage
\bibliographystyle{splncs}  
\bibliography{biblio}
\end{document}